\documentclass[lettersize,journal]{IEEEtran}
\usepackage{amsmath,amsfonts,amssymb}
\usepackage{algorithmic}
\usepackage{algorithm}
\usepackage{array}
\usepackage{textcomp}
\usepackage{color}
\usepackage{stfloats}
\usepackage{url}
\usepackage{verbatim}
\usepackage{graphicx}
\usepackage{cite}
\usepackage{graphicx}
\usepackage{amsmath}
\usepackage{amssymb}
\usepackage{subfigure}
\usepackage{multirow}
\usepackage{tabularx}
\usepackage{booktabs}
\usepackage{diagbox}
\usepackage{paralist}
\usepackage{epstopdf}
\usepackage{color}
\def \ie {{i.e.}~}
\def \eg {{e.g.}~}
\def \etc {{etc.}~}
\def \etal {{et al.}~}
\def \vs {{v.s.}~}
\def \wrt {{w.r.t.}~}

\newcommand{\mshi}[1]{#1}
\newcommand{\miaojing}[1]{#1}

\newcommand{\para}[1]{\noindent \textbf{#1}}

\begin{document}

\title{MFNet: Multi-class Few-shot Segmentation Network with Pixel-wise Metric Learning}

\author{Miao Zhang,
        Miaojing Shi,~
        and~Li Li
\thanks{Miaojing Shi and Li Li are corresponding authors.}
\thanks{Miao Zhang and Li Li is with the  College of Electronic and Information Engineering, Tongji University. E-mail: zhangmiao1997@tongji.edu.cn, lili@tongji.edu.cn.}
\thanks{Miaojing Shi is with the Department of Informatics, King's College London. E-mail: miaojing.shi@kcl.ac.uk}
}

\markboth{Accepted on IEEE Transactions on Circuits and Systems for Video Technology}%
{Zhang \MakeLowercase{\textit{et al.}}}


\maketitle
\makeatletter

\begin{abstract}
In visual recognition tasks, few-shot learning requires the ability to learn object categories with few support examples. Its re-popularity in light of the deep learning development is mainly in image classification. This work focuses on few-shot semantic segmentation, which is still a largely unexplored field. A few recent advances are often restricted to single-class few-shot segmentation. In this paper, we first present a novel multi-way (class) encoding and decoding architecture which  effectively fuses multi-scale  query  information and  multi-class support information into one query-support embedding. Multi-class segmentation is directly decoded upon this embedding. For better feature fusion, a multi-level attention  mechanism is proposed within the architecture, which includes the attention for support feature modulation and attention for multi-scale combination. Last, to enhance the embedding space learning, an additional pixel-wise metric learning module is introduced with triplet loss formulated on the pixel-level of the query-support embedding. 
Extensive experiments on standard benchmarks PASCAL-$5^{i}$ and COCO-$20^{i}$ show clear benefits of our method over the state of the art in \miaojing{multi-class} few-shot segmentation. Our codes will be available at \emph{\color{magenta}{https://github.com/roywithfiringblade/MFNet}}.
\end{abstract}

\begin{IEEEkeywords}
few-shot segmentation, multi-class, attention, metric learning
\end{IEEEkeywords}

\section{Introduction}\label{sec:intro}
\IEEEPARstart{D}{eep} learning techniques have been widely employed for a broader spectrum of computer vision applications ranging from image classification~\cite{simonyan2015iclr,he2016cvpr}, object detection~\cite{ren2016pami,he2017iccv} and semantic segmentation~\cite{long2015cvpr,chen2017pami}. With plenty of data and advanced hardware, great success has been achieved in these tasks using deep neural networks (DNNs). These networks can be good experts for specific tasks they are trained with, notwithstanding, easily become novices when facing new tasks with few examples.
{To address this challenge, few-shot learning is proposed. Unlike conventional methods trained with a large amount of data, it requires only a small number of examples. Representative works~\cite{vinyals2016arxiv,snell2017arxiv} extract prototype representations from few labeled data per class and use them to match the object class of query data. }

Many works in few-shot learning are for image classification~{\cite{tokmakov2019cvpr,jiang2020multi,toan,Hao2021csvt}}, with a few applied to object detection~\cite{kang2019iccv,yang2020nips,Cheng2021csvt,yang2021training}, semantic segmentation~\cite{zhang2019cvpr,wang2019iccv,liu2020eccv,Liu2020DynamicEN,zhang2021few}, \etc This work focuses on few-shot semantic segmentation (FSS): we are given a few pixel-wisely annotated support images to perform semantic segmentation on new classes. The DNN is trained with episodes, each containing a handful of support images and a query image. It learns class information from support images and performs pixel-level classification on the query image.  The prediction is optimized with the ground truth query mask as the supervision.
For extracting knowledge from support images, many popular cues such as attention~\cite{zhang2019cvpr}, context~\cite{liu2020eccv}, and multi-scale~\cite{tian2020pami} have been employed. For pixel-level classification on the query, it is more complicated than image-level classification and is a key challenge.

\begin{figure}
    \centering
    \includegraphics[scale = 0.63]{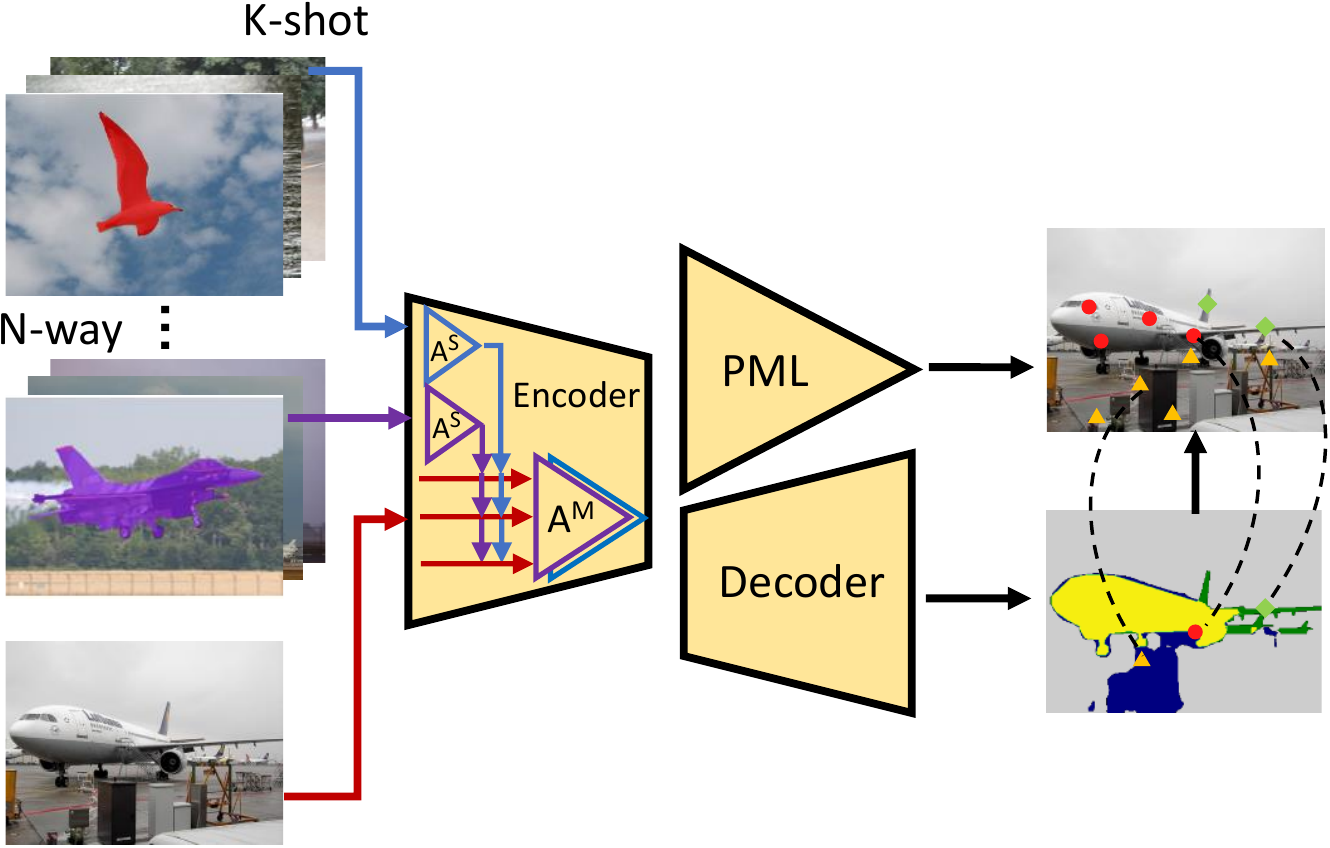}
    \caption{Our proposed multi-class few-shot (N-way K-shot) segmentation network (MFNet). It offers a new multi-way encoding and decoding architecture equipped with a multi-level attention mechanism for better modulation of support features (A$^\text{S}$) and combination of multi-scale features (A$^\text{M}$).  A pixel-wise metric learning (PML) branch is also included for embedding space optimization. }
    \label{fig:intro}
\end{figure}

To solve the per-pixel classification in FSS, 
different ways are introduced to fuse feature representations of the query image and support images in the network and apply pixels-to-pixels predictions~\cite{zhang2019cvpr,zhang2019iccv,tian2020pami,liu2020cvpr,Zhang2021SelfGuidedAC,lu2021iccv}. These works, as indicated by~\cite{tian2020aaai}, are mostly designed for single-class segmentation and cannot be easily extended to multi-class segmentation.
\mshi{However, in the real world, most photos/videos that we face contain multi-class information. Multi-class task is indeed a more practical task than  single-class task. As a matter of fact, many popular computer vision datasets, \eg PASCAL VOC~\cite{everingham2010ijcv} and MS-COCO~\cite{lin2014eccv}, were curated to contain multi-class information; and many established computer vision algorithms, \eg Faster R-CNN~\cite{ren2016pami}, Mask R-CNN~\cite{he2017iccv}, and FCN~\cite{long2015cvpr}, were also developed to solve multi-class problems. Hence, solving the FSS in a multi-class scenario is both realistic and valuable.}

There are some other works in FSS who conduct the per-pixel classification via explicit similarity measure between features of the query image and support images~\cite{wang2019iccv,dong2018bmvc,liu2020eccv,tian2020aaai}.
They can handle multi-class segmentation with multiple classifiers. Notwithstanding, the classifiers mainly rely on support images while the query information is not fully exploited. More importantly, the per-pixel classification is based on global descriptors of classes which may have noise because of intra-object variation~\cite{zhang2019iccv}.

This paper aims to design a novel end-to-end, pixels-to-pixels DNN for multi-class few-shot semantic segmentation (MFNet, Fig.~\ref{fig:intro}).
\IEEEpubidadjcol
First, a novel multi-way (class) encoding and decoding architecture is introduced where we extract both multi-scale query features and multi-class prototype features from the network by feeding into the query and support images.
In order to obtain effective class prototypes from multiple support images, an attention scheme is introduced to extract relational features within the support set of each class and use them to modulate the original features of support images. The modulated support features are averaged to produce the class prototype. Each class prototype is pixel-wisely encoded into the multi-scale query features, respectively. They are combined via another attention scheme which utilizes self-attended weights on each scale for the fusion. It results into one feature per class.
Multiple features over different classes are further concatenated to create one query-support embedding.
A parametric decoder is directly applied on it to predict the pixel-level multi-class probabilities on the query. 
In addition, to enhance the query-support embedding space learning, we introduce a pixel-wise metric learning (PML) module with triplet loss~\cite{schroff2015cvpr}. The triplet is formulated on the query's pixel-level where hard positive/negative pairs are chosen based on the false negative/positive pixels \wrt the ground truth (green and blue pixels in Fig.~\ref{fig:intro}: output). A weighted random selection strategy is introduced to select positive pairs depending on the spatial distance between pixels. 


To summarize, the contribution of this work is three-fold:

\begin{itemize}
   \item We introduce a novel multi-way encoding and decoding architecture, which effectively fuses multi-scale query information and multi-class support information and
   is the first of its kind.
   \item For better feature fusion, we propose a multi-level attention mechanism, which includes two different attention schemes for support feature modulation and multi-scale combination, respectively.
\item  To enhance embedding space learning, we introduce an pixel-wise metric learning module, which defines triplet loss on the pixel-level query-support embedding.

\end{itemize}

Extensive experiments on standard benchmarks PASCAL-$5^{i}$~\cite{shaban2017bmvc} and  COCO-$20^{i}$~\cite{wang2019iccv,nguyen2019iccv}  demonstrate that our method significantly improves the state of the art FSS solutions, especially in the multi-way setting.

\section{Related Work}\label{Sec:relatedwork}
\subsection{Semantic segmentation}
The task of semantic segmentation is to predict the per-pixel label of an image.  Long \etal~\cite{long2015cvpr} first designed a fully convolution network which significantly advances the research in this field.
Subsequent works such as DeepLab~\cite{chen2017pami}, DPN~\cite{liu2015iccv}, EncNet~\cite{zhang2018cvpr} improve the segmentation performance by integrating MRF/CRF or contextual information into the pipeline.
\miaojing{Recently, Ji \etal~\cite{Ji2021csvt} proposed a cascaded CRFs to learn the object boundary information from multiple layers of the neural network, which helps to produce more accurate segmentation result. Weng \etal~\cite{XI2021csvt} introduced a stage-aware feature alignment module to perform real-time semantic segmentation of street scenes, which aligns and aggregates different levels of feature maps effectively. Nirkin \etal~\cite{Nirkin2021HyperSegPH} designed a hypernetwork for the encoder-decoder architecture that can generate weights for the decoder to enable a fast and accurate segmentation prediction. Different from CNN-based methods, Strudel \etal~\cite{Strudel2021SegmenterTF} designed a transformer-based model which can capture the global context from the first layer of the network and throughout the encoding and decoding stages. The model yields state-of-the-art results on several standard benchmarks.}
The success of these works relies on the large amount of training images and elaborate annotations on image pixels. Collecting such a large corpus can be tedious and time consuming. Therefore, many recent works resort to weakly/semi-supervised learning{~\cite{chen2020eccv,mittal2019semi,meng2019weakly,wang2022semi}} as well as few-shot learning{~\cite{zhang2019cvpr,Pambala2021SMLSM,wang2019iccv,liu2020eccv}}. 

\subsection{Few-shot learning}
{Few-shot learning refers to learning from a few labeled samples, which has become re-popularized recently with the advent of deep learning and is mainly adopted in image classification. Meta-learning and metric-learning are frequently used techniques in this field~\cite{koch2015icmlw,vinyals2016arxiv,snell2017arxiv,jiang2020multi,toan,Hao2021csvt}. For instance,  Koch ~\etal ~\cite{koch2015icmlw} proposed a Siamese framework with twin branches taking input of query and support images respectively, then measure their distance by the end of the network. Vinyals~\etal~\cite{vinyals2016arxiv} employed a similar structure like ~\cite{koch2015icmlw} but utilized an LSTM to get a context embedding of support images.~\miaojing{Jiang ~\etal ~\cite{jiang2020multi} designed a multi-scale metric learning method to extract multi-scale features and mine their relations for the few-shot classification. Huang~\etal~\cite{toan} presented a target-oriented alignment network to reduce the intra-class variance and enhance the inter-class discrimination in the  fine-grained classification problem. Hao~\etal~\cite{Hao2021csvt} designed a semantic-alignment based global-local interplay metric learning framework, which focuses on learning the semantics that are relevant to the global information extracted from the query-support image pair.
}
There are also some methods utilizing data augmentation techniques~\cite{chen2018arxiv,wang2018cvpr} to {cope} with the lack of training samples in few shot learning.}

\miaojing{Above works are mostly for the image classification/recognition task, while in this paper we study the few-shot semantic segmentation task, specified below. }


\subsection{Few-shot semantic segmentation}
Few-shot semantic segmentation differs from image classification as it requires per-pixel classification~\cite{dong2018bmvc,zhang2020tc,zhang2019cvpr,tian2020pami,liu2020cvpr,Pambala2021SMLSM}. Early works tend to adapt techniques from few-shot image classification to segmentation~\cite{shaban2017bmvc,dong2018bmvc,zhang2020tc}. For instance,
Zhang~\etal~\cite{zhang2020tc} utilized the masked average pooling to generate class prototypes and use them to classify pixels on the query via cosine similarity.
{Recently, Wang \etal~\cite{wang2019iccv} designed a novel prototype alignment regularization module, which lets the support image and query image predict the result mutually for each other. Tian~\etal~\cite{tian2020pami} extracted prior information from pre-trained backbone and proposed a prior guided feature enrichment network to optimize the segmentation using multi-scale features.
\miaojing{Pambala~\etal~\cite{Pambala2021SMLSM} designed a meta-learning framework, which integrates the attribute information with the visual feature embeddings to make the class prototype more accurate and robust in few-shot segmentation.} Li~\etal~\cite{Li2021AdaptivePL} introduced an adaptive superpixel-guided network to match the query and support features effectively with the use of superpixel clustering and prototype refining. Lu ~\etal~\cite{lu2021iccv} introduced a classifier weight transformer to adapt the weight of the  classifier in the segmentation network according to different query images. \miaojing{ Min~\etal~\cite{min2021hypercorrelation} leveraged multi-level feature correlation and 4D convolutions to extract diverse features from different levels of convolutional layers, which is proved to be effective in fine-grained segmentation with limited supervision.}}

\mshi{Most works in FSS only work in the single-class setting while the datasets they work on were originally curated for multiple classes. For instance, there are on average 3.5 labelled object classes per image in the MS-COCO dataset~\cite{lin2014eccv}. Moreover, similar tasks such as few-shot classification and few-shot object detection are also considered as the multi-class problem~\cite{vinyals2016arxiv,jiang2020multi,kang2019iccv,Cheng2021csvt}. To bridge the gap between FSS and other few-shot learning tasks,  we introduce a novel end-to-end, pixels-to-pixels DNN for multi-class few-shot segmentation.}

\para{Architecture.} Works in~\cite{zhang2019cvpr,tian2020pami,liu2020cvpr} fuse features of query and support images in the network and directly decode segmentation results upon the fused feature. They only report single-class segmentation. \cite{tian2020aaai,Liu2020DynamicEN} explicitly learn  classifiers from support images to predict the label of each query pixel for multi-class segmentation. 
Others like~\cite{wang2019iccv,dong2018bmvc,liu2020eccv} employ non-parametric classifiers (e.g. nearest neighbor) in their models which are extendable to multi-class but need to be computed for multiple times. 
Our MFNet is similar to~\cite{zhang2019cvpr,tian2020pami,liu2020cvpr} to directly decode segmentation results from latent features. Unlike them, we design a first end-to-end pixels-to-pixels architecture for multi-class few-shot segmentation.
Moreover, we propose a new multi-level attention mechanism for better support feature modulation and multi-scale combination; a new pixel-wise metric learning module to optimize the embedding space for multi-class segmentation.
\begin{figure*}[htp]
    \centering
    \includegraphics[scale=0.63]{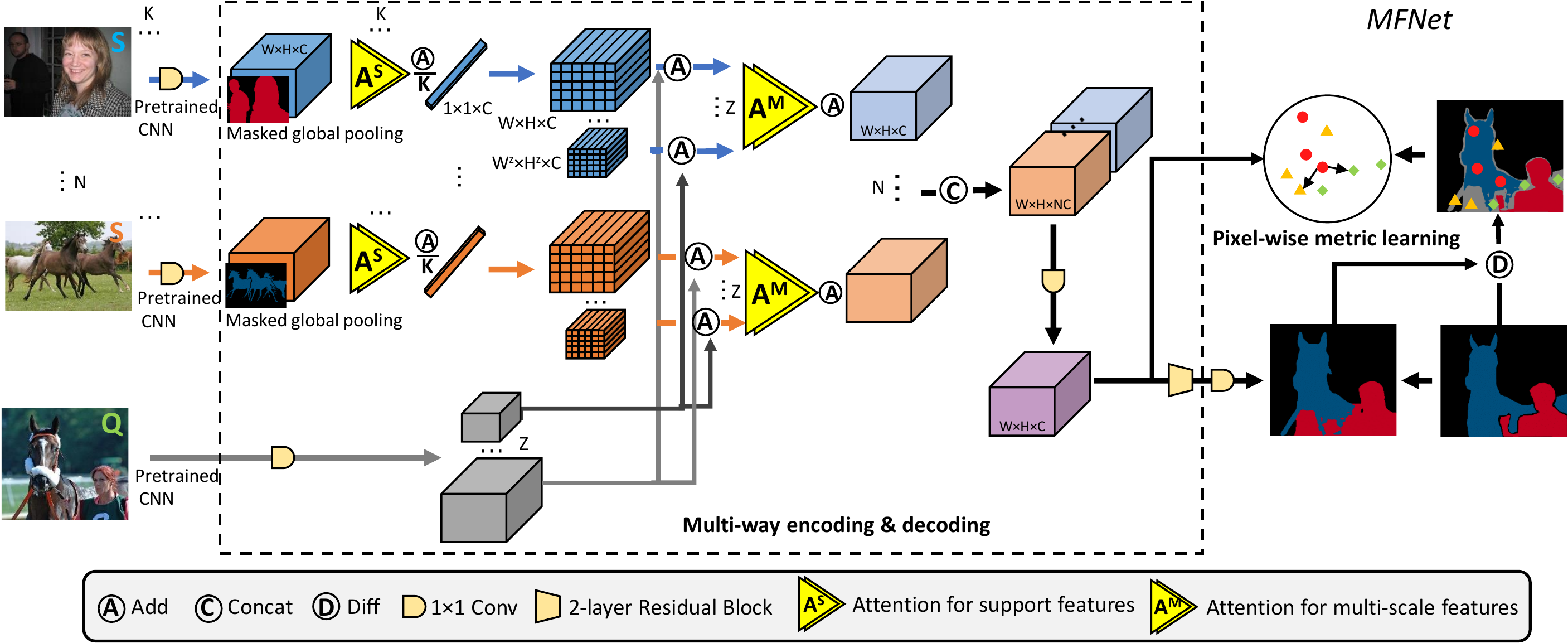}
    \caption{MFNet pipeline: features of support images ($S$) and multi-scale ($Z>1$) features of the query image ($Q$) are extracted via a pretrained CNN. Masked global pooling is applied to support features to convert them into $1 \times 1 \times C$ tensors. An attention scheme A$^\text{S}$ based on feature relations is proposed to modulate multiple support features ($K > 1$) within each class and then average them  into one class prototype. The class prototype is expanded to the same shape of a query feature of certain scale and added with it. These merged multi-scale features are combined with another attention scheme A$^\text{M}$ (shareable for $N$ branches of classes) which computes self-attended weights on each scale. It results into one feature per class. We concatenate features over $N$ classes into one query-support embedding and let the network directly decode multi-class segmentation on it. A pixel-wise metric learning module is branched off in the decoder. Triplets are selected based on the pixel spatial distances and label differences between predictions and ground truth.  A$^\text{S}$ and  A$^\text{M}$ are specified in Fig.~3.
    }
    \label{fig:overview}
\end{figure*}

\para{Attention.} Many works have utilized the attention mechanism in few-shot segmentation~\cite{zhang2019cvpr,hu2019aaai,liu2020eccv,Liu2020DynamicEN,azad2021wacv,yang2020bmvc}. Common practices include attention within images~\cite{hu2019aaai,zhang2019cvpr} and across images~\cite{liu2020eccv,Liu2020DynamicEN,azad2021wacv,yang2020bmvc,wu2021iccv}. The latter can be implemented among support images{~\cite{liu2020eccv,Liu2020DynamicEN}} or between query and support images~\cite{zhang2019cvpr,azad2021wacv,yang2020bmvc,wu2021iccv}. Building upon these achievements, this paper proposes a multi-level attention mechanism which includes two different attention schemes for support feature modulation and multi-scale combination, respectively.
The first attention computes relational features among support images to modulate original support features, which is motivated by~\cite{hu2018cvpr} from object detection. \mshi{The relational module in ~\cite{hu2018cvpr} includes a geometry weight to encode spatial information among objects in an image while we do not need this as we compute the relation of support features across images; the relation in ~\cite{hu2018cvpr} is calculated across object classes while we compute the intra-class support feature relation.}
Our attention also differs from~\cite{liu2020eccv} who clusters support features into multiple prototypes each class and uses a graph attention network to modulate them. The whole process in~\cite{liu2020eccv} concerns both labeled and unlabeled data, online message passing and refinement, which can not be easily deployed off-the-shelf; ours is simpler and more effective. The second attention takes the inspiration from~\cite{Chen2016cvpr} to fuse multi-scale outputs via attended weights. It differs from~\cite{zhang2019cvpr} whose attention is computed on each support image and module is also constructed differently.

\para{Metric learning.} The embedding space in our network is reinforced via a distance metric learning module with pixel-level triplet loss. \mshi{Triplet loss dates back to~\cite{weinberger2005distance} where the essential idea is to push samples of the same class close and pull samples of different classes away in the learning space. Schroff~\etal~\cite{schroff2015cvpr} have adapted it in the face recognition task and it has been re-popularized since then. Many following works focus on proposing new sampling strategies and definitions of feature distance in the loss function~\cite{karlinsky2019cvpr,Tian2020HyNetLL,Zeng2020HierarchicalCW}.} Triplet loss has been employed in few-shot learning before but is mostly on image/region-level~\cite{wang2018arxiv,karlinsky2019cvpr}. Pixel-level loss has been utilized in other tasks~\cite{chen2018cvpr,qian2019aaai,zhang2020nips,van2022dam} but not in the few-shot setting. Our triplet sampling strategy differs from~\cite{chen2018cvpr,karlinsky2019cvpr,Tian2020HyNetLL,Zeng2020HierarchicalCW,qian2019aaai,zhang2020nips,schroff2015cvpr}: we introduce a weighted random selection strategy to select pixel pairs based on their spatial distances as well as corresponding predictions by the network. It is fast and effective.

\section{Method}\label{Sec:method}
\subsection{Problem setting}\label{Sec:prob-setting}
We study the few-shot semantic segmentation where the model is trained on a set of base (seen) classes $\mathcal C_{tr}$ for semantic segmentation, and is expected to perform fast segmentation on a set of novel (unseen) classes $\mathcal C_{te}$ ($\mathcal C_{tr} \cap \mathcal C_{te} = \varnothing $) with only few annotated support images.

We perform $N$-way (class) $K$-shot semantic segmentation. Both training and testing are organized in episodes: an $N$-way $K$-shot episode is an instance of the few-shot
task represented by a support set $\mathcal S$ of $K$ training images from
each of the $N$ classes, and a query image of object(s) belonging to the $N$ classes. We denote by $\mathcal S_{n} = \{S_{n1}, S_{n2},...,S_{nK}\}$ the support set for the $n$-th class and $Q$ the query image. For each support image $S_{nk}$, it is also associated with a ground truth label mask $M_{nk}$. The ground truth for the query image is also provided, which is used for network optimization during training.

\subsection{Multi-way encoding and decoding architecture} \label{Sec:MED}
An overview of our MFNet is in Fig.~\ref{fig:overview}: the backbone design follows~\cite{zhang2019cvpr,tian2020pami}, \ie extracting the middle-level features (conv3$\_$x and conv4$\_$x) from ResNet-50. We use $\Phi$ to denote this feature extractor.
Given a query image and multiple support images from different classes, they are firstly fed into the shared backbone to extract features. In particular, for the query $Q$, we follow~\cite{tian2020pami} to extract multi-scale query features \{$\Phi^z(Q)$\} using adaptive average pooling ($Z$ = $4$ in this paper), they are of size $W^z \times H^z \times C$, respectively. This enrichment strengthens the interaction between query and support features. For the $k$-th support image in the $n$-th class, $S_{nk}$, we have its feature $\Phi(S_{nk})$. Masked global pooling is applied to $\Phi(S_{nk})$ such that its per-pixel features in the background are filtered out via $M_{nk}$ and in the foreground are averaged into a $1 \times 1 \times C$ tensor $F_{nk}$. The class prototype of $n$-th class is obtained by averaging $F_{nk}$ over $K$-shot images: $F_n = \frac{1}{K}\sum_k^K F_{nk}$. For $K > 1$, a attention scheme based on feature relations is introduced to modulate the $K$-shot features before their average (specified later). $F_n$ is of size $1 \times 1 \times C$ while $\Phi^z(Q)$ is of size $W^z \times H^z \times C$. To merge $F_n$ with \{$\Phi^z(Q)$\}, we replicate the $F_n$ by $W^z \times H^z$ times and pixel-wisely add them to each $\Phi^z (Q)$, respectively (see Fig.~\ref{fig:overview}).
These merged multi-scale features, denoted as \{$X_n^z$\} for the $n$-th class, are combined via self-attended weights on each scale (specified later), which result into one feature tensor $S_n$ ($W \times H \times C$) for this class. For $N$ classes, we concatenate $N$ feature tensors as one query-support embedding of size $W \times H \times NC$. 

The feature fusion between the query and each class feature uses ``Add" operation while among different classes uses ``Concat". Given two features, if concatenated, they shall be processed with different weights. This suits the aim for the latter fusion among different classes. While for the former fusion, we choose ``Add" to process the query and class feature equivalently as a whole, which empirically works better than ``Concat". Having a combinatorial view, the former does not use ``Concat" is also to avoid the redundancy of query features brought over different classes in the next ``Concat". 
The query-support embedding contains both the query and class information. We can decode per-pixel class probabilities on it. The decoder is of a $1\times 1$ conv, a residual block, and another $1\times 1$ conv (Fig.~\ref{fig:overview}). The first conv reduces the number of embedding channels back to $C$. The residual block remains with $C$ channels and the last conv reduces the number of channels to $N+1$ for multi-class (including background) prediction.
Let $p_{mn}$ be the probability for $n$-th class at $m$-th pixel, we utilize a weighted focal loss to supervise the segmentation result:
\begin{equation}
\begin{split}
     &\mathcal L_\text{SEG}  =-\frac{1}{MN}\sum_{m}^M\sum_{n}^N \omega_n{(1-p_{mn})}^{\gamma}y_{mn}\log(p_{mn}) \\
     &\omega_n = 1/\log(1.1+M_n/M),
\end{split}\label{eq:seg}
\end{equation}
\noindent where $y_{mn}$ is the ground truth label: if the $m$-th pixel belongs to the $n$-th class, $y_{mn} = 1$; otherwise, $y_{mn} = 0$. $M$ is the total number of pixels and $M_n$ is the total number of pixels belonging to class $n$. $\omega_n$ is a weighting factor which adjusts the class influence in (\ref{eq:seg}) depending on the ratio of $M_n$ and $M$. We take the form of $\omega_n$ to give a small value of it if $M_n/M$ is big and vice versa.

\begin{figure*}[htbp]
\centering
\subfigure[]{
\begin{minipage}[t]{0.5\linewidth}
\centering
\includegraphics[scale=0.32]{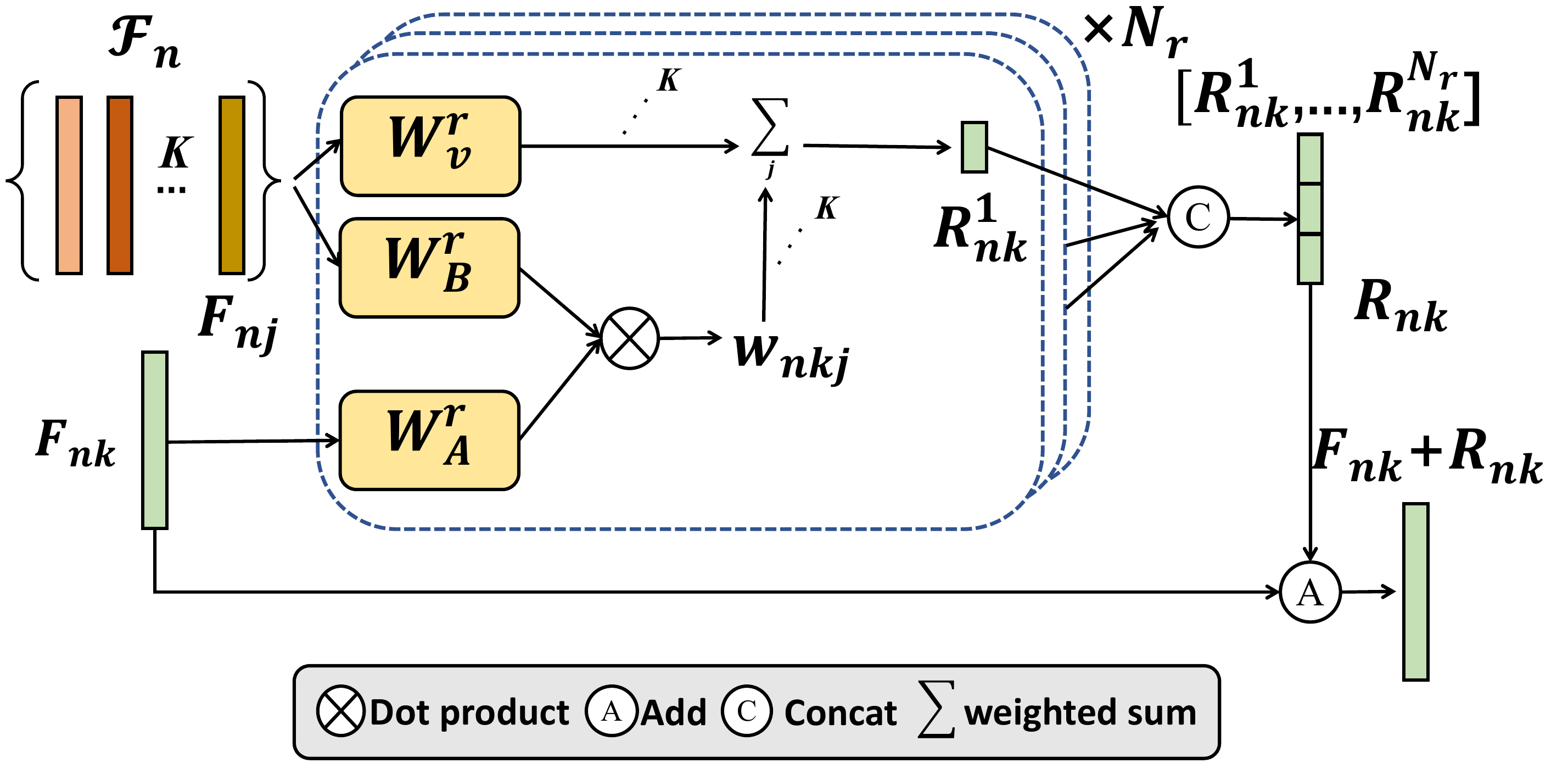}
\label{As}
\end{minipage}%
}%
\subfigure[]{
\begin{minipage}[t]{0.5\linewidth}
\centering
\includegraphics[scale=0.32]{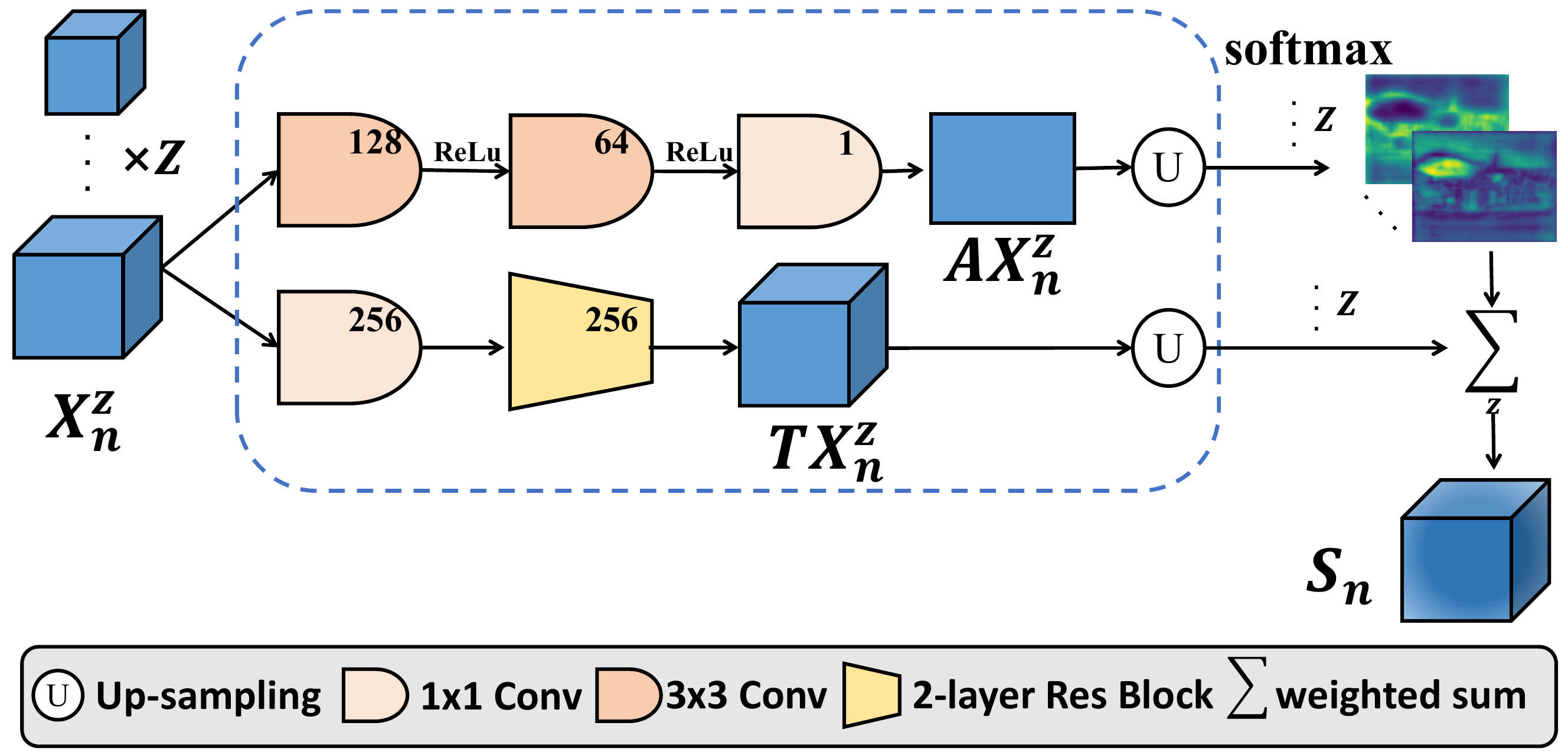}
\label{Am}
\end{minipage}%
}%
\centering
\caption{\miaojing{Illustration of the attention module $A^\text{S}$ and $A^\text{M}$ for support feature modulation (a) and multi-scale combination (b), respectively.}}
\end{figure*}
\subsection{Multi-level attention}
We introduce a multi-level attention mechanism which includes two different attention schemes for better modulation of multiple support features and better combination of multi-scale features, respectively. They are of different purposes: \miaojing{on the support feature level, given an object class, its appearance and context vary across different support images. To tackle this intra-class variance, we utilize the relational features to extract common traits among multiple support features to represent the class.
On the other hand, multi-scale features fire on different levels of details for a certain object class. Their combination in previous works, \eg the FEM module in~\cite{tian2020pami}, normally gives equal weight to each scale. To capture the distinctive strength of each scale feature,
we compute the self-attention on each scale and utilize the attention maps to perform a weighted sum of multi-scale features.
Both attention schemes devise shareable modules among $N$ branches of the classes. }

\noindent \para{Attention for support feature modulation.} In order to efficiently merge features of support images in the K-shot setting, inspired by~\cite{hu2018cvpr}, we introduce an attention scheme based on support feature relations {(see Fig.~\ref{As})}. Given the support set of $\mathcal S_n = \{S_{nk}\}$ for class $n$, by extracting their features and applying the masked global pooling, we obtain the corresponding set of embedding vectors $\mathcal F_n = \{F_{nk}\}$. The relational feature $R_{nk}$ of the whole $\mathcal F_n$  with respect to the $k$-th support feature $F_{nk}$ is computed as
\begin{equation}
    R_{nk} = \sum_{j=1}^{K} {w}_{nkj} \cdot (W_V\cdot F_{nj})
    \label{eq:att}
\end{equation}
where $W_V$ is a transformation matrix; the relation weight ${w}_{nkj} $ is computed as the softmax of the dot product between the transformed feature $W_A F_{nk}$ and $W_B F_{nj}$:
\begin{align}
    {w}_{nkj} &= \text{softmax} (\frac{\text{dot}(W_A F_{nk}, W_B F_{nj})}{\sqrt{d_k}})
\end{align}
$W_A$ and $W_B$ are matrices that project $F_{nk}$ and $F_{nj}$ to the dimension of $d_k$, respectively. 
\miaojing{Like in~\cite{hu2018cvpr}, we can compute $N_r$ relational features ($R_{nk}^1$,…,$R_{nk}^{N_r}$) for $F_{nk}$ such that the channel of each is $1/N_r$ of the dimension of $F_{nk}$ (the output channels of $W_A^r$, $W_B^r$, $W_V^r$ are also set to $1/N_r$ of that of $F_{nk}$). $R_{nk}^1$,…,$R_{nk}^{N_r}$ are concatenated to form $R_{nk} = \text{Concat} [R_{nk}^1,...,R_{nk}^{N_r}]$, which serves as a  modulator to $F_{nk}$ and is added to it. }

The modulated support features are further averaged to create a representative prototype of this class.

\medskip

\noindent \para{Attention for multi-scale combination. }
The multi-scale features pay attention to different levels of details. For better combination of them, we extract the attention on each scale. Attentions are normalized over multiple scales and multiplied to scale features, respectively. These attended features are added together such that levels of details over multiple scales are selectively incorporated into one feature tensor.

\miaojing{Specifically, 
given the $z$-th scale’s feature $X_n^z$ of size ${W^z\times H^z\times C}$ for the $n$-th class, we illustrate its self-attention in Fig.~\ref{Am}. It consists of two branches: in the upper branch, $X_n^z$ is passed through two 3$\times$3 and one 1$\times$1 convolutional layers to obtain the attention map $AX_n^z$ of size $W^z\times H^z\times 1$; in the lower branch, $X_n^z$ is passed through one 1$\times$1 convolutional layer and one 2-layer residual block to obtain the transformed feature $TX_n^z$ ($W^z\times H^z\times C$). Since multi-scale features have different resolutions, we up-sample $AX_n^z$ and $TX_n^z$ to the highest resolution over scales, \ie $W\times H\times 1$ and $W\times H\times C$, respectively. The upsampled $AX_n^z$, which we denote as U$(AX_n^z)$, is per-pixel softmaxed over other attention maps from different scales. \mshi{We use the normalized attention maps as element-wise weights, \ie $a_n^z=softmax(\text{U}(AX_{n}^{z}))$,} to multiply their corresponding upsampled scale features U$(TX_n^z)$, so as to perform a weighted sum of the multi-scale features:}
\begin{align}
    a_n^z&=softmax(\text{U}(AX_{n}^{z}))\\
    S_n&=\sum_{z=1}^{Z} \text{U}(TX_{n}^{z})\cdot a_n^z
\end{align}
\miaojing{$S_n$ from different classes are further concatenated to produce the final query-support embedding. }

\miaojing{This attention scheme can be particularly beneficial in the multi-way setting: because different object classes may appear in different positions and be with different sizes in one image, often surrounded by different local contexts. The proposed attention scheme carefully takes care of this inter-class variance by producing tailored weight maps to combine multi-scale features for every object class.  }

\subsection{Pixel-wise metric learning } \label{Sec:PML}
To further enhance the embedding space learning for the query and support images, we employ the metric learning which commonly adopts contrastive loss~\cite{chopra2005cvpr,hadsell2006cvpr} or triplet loss~\cite{schroff2015cvpr} to learn effective feature representations for similar images/patches. \miaojing{The metric learning in~\cite{chopra2005cvpr,hadsell2006cvpr,schroff2015cvpr} is operated on the image level for visual recognition task.} In this work, we are keen on the accuracy of pixel-wise classification and therefore formulate the metric learning on the pixel-level query-support embedding. Given the segmentation prediction by the network, we can obtain its false positives/negatives with respect to the ground truth. For a specific class, false positives are false pixels being mis-classified as this class which can be seen as hard negatives. False negatives are true pixels being mis-classified as not this class which can be seen as hard positives. Anchors can be selected from the correctly classified pixels of this class. This forms three pools of hard positives, hard negatives, and anchors, respectively. We can write out a triplet loss function by randomly selecting elements from them. This works but has demerits: due to the intra-object variation among pixels, embedding vectors of pixels can vary a lot in different parts of an object. Intuitively, object pixels within a small neighborhood are more likely to have similar embedding vectors. We therefore propose a weighted random selection strategy to select pixel pairs depending on their spatial distances. 

We denote by $\mathcal A$, $\mathcal {HP}$ and $\mathcal {HN}$ the above obtained pools of pixels for anchors, hard positives, and hard negatives, respectively. To construct a triplet: first, a pixel $a$ is randomly selected from $\mathcal A$, where $a$ is associated with its spatial coordinates ($x_a$, $y_a$) in the image as well as its embedding vector $f_a$ in the network. Next we select the hard positive pixel from $\mathcal {HP}$ to form a pair with $a$. Based on the discussion about the intra-object variation, we compute the spatial distance from the coordinates ($x_a$, $y_a$) of $a$ to that ($x_{hp}$, $y_{hp}$) of every candidate $hp$ in $\mathcal {HP}$. The distance is converted into probabilities for a weighted random selection such that the pixel $hp^*$ who is spatially closer to $a$ is more likely to be selected; $\langle a, hp^* \rangle$ is a meaningful hard positive pair. As for the hard negative $hn^*$ to $a$, a similar spatial constraint might apply among adjacent pixels.
Yet, this adjacency is not always observable in $\mathcal A$ and $\mathcal {HN}$ because in reality the small neighborhood of anchors are most likely to be positives. This is to say, hard negatives do not have direct relations to their spatial distances to anchors. A random selection from $\mathcal {HN}$ is sufficient. Having the triplet $\langle a, hp^*, hn^* \rangle$ and their feature embedding $\langle f_a, f_{hp^*}, f_{hn^*} \rangle$, the triplet loss is written as,
\begin{equation} \label{eq:triplet}
    \mathcal L_\text{PML} = \sum_{\langle a, hp^*, hn^* \rangle} \max(\|f_a - f_{hp^*} \|_2^2 -  \|f_a - f_{hn^*} \|_2^2 + \alpha, 0)
\end{equation}
We do not elaborate every pixels in $\mathcal A$, $\mathcal {HP}$ and $\mathcal {HN}$ but only a subset of them (\ie $N_t$ seeds in $\mathcal A$). The pixel feature embedding is taken before the residual block in the decoder.

Notice 1) for one image containing multiple classes, $N_t$ initial seeds are randomly selected from the joint pools of anchors of these classes, triplets are then formed in corresponding pools of each class; 2) it is also possible to find the positive/negative pairs via the maximum/minimal feature distances among pixels~\cite{schroff2015cvpr}. We did not choose this because it is computationally expensive to compute feature distances among all foreground and background pixels online. In contrast, the spatial distances can actually  be computed offline.

The overall loss function is given by
\begin{equation}\label{eq:loss}
   \mathcal L = \mathcal L_\text{SEG} +  \lambda \mathcal L_\text{PML}.
\end{equation}
$\lambda$ is a loss weight.

\subsection{Inference}
At testing stage, when it comes with query images for novel classes and a few annotated support images of these classes, our MFNet does not need fine-tuning. Instead, we pass certain query image and those support images of different novel classes through MFNet to obtain their embeddings, which are further fused in the network to directly predict the segmentation result on the query image.

\section{Experiments}
\subsection{Datasets}
Following previous works~\cite{shaban2017bmvc,wang2019iccv,nguyen2019iccv} we conduct experiments on two commonly used few-shot semantic segmentation datasets: PASCAL-$5^{i}$   and COCO-$20^{i}$.

\para{PASCAL-$5^{i}$} is constructed from PASCAL VOC 2012~\cite{everingham2010ijcv} with SBD augmentation~\cite{hariharan2011iccv}. 20 classes from the original dataset are divide into 4 folds, each containing 5 classes. Models are trained on three folds and tested on the other fold. Like~\cite{wang2019iccv}, we repeat each experiment five runs and report the average results. Each run has 1,000 episodes. 

\para{COCO-$20^{i}$} is built from MS COCO~\cite{lin2014eccv}. Compared to PASCAL-$5^{i}$, it is with more object classes. 80 classes are divided into 4 folds, each containing 20 classes.
Models are trained on 60 classes and tested on the rest 20 classes. We follow the practice in PASCAL-$5^{i}$ to repeat each experiment five runs, with each run containing 10,000 episodes.

\subsection{Implementation details and evaluation protocol}\label{sec:implement}
\para{Implementation details.} Similar to~\cite{tian2020pami,zhang2019cvpr}, we adopt the backbone from ResNet-50 pretrained on the ILSVRC classification task. Backbone weights are fixed during the training. Input images are resized to 473 $\times$ 473. Random crop, rotation and mirror operation are applied for data augmentation. $W$, $H$, and $C$ in Sec.~\ref{Sec:MED} are 60, 60, and 256, respectively. $\alpha$ in (\ref{eq:triplet}) is set to 1 while $\lambda$ in (\ref{eq:loss}) is set to 0.4. $N_t$ for triplet is 20 and $N_r$ for relational attention is 4. The network is trained by the SGD optimizer with a momentum of 0.9 and a weight decay of 0.0001; the `poly' policy is adopted to decay the learning rate with a power of 0.9. We train the network for 200 (50) epochs on PASCAL-$5^{i}$ (COCO-$20^{i}$)  with an initial learning rate of 0.0025 (0.005) and a batch size of 4 (8). The PML loss $\mathcal L_\text{PML}$ is utilized after 5 epochs on PASCAL-$5^{i}$ and 1 epoch on COCO-$20^{i}$. \mshi{All experiments are conducted on the NVIDIA GeForce RTX 2080 Ti GPU.}

Our train/test setup by default follows~\cite{wang2019iccv,liu2020eccv} where on PASCAL-$5^{i}$ we run the 2-way 1-shot and 2-way 5-shot experiments while on COCO-$20^{i}$ we run the 2-way 1-shot and 5-way 1-shot experiments. The 1-way 1-shot and 1-way 5-shot experiments are also conducted on these datasets.
Notice 1) in the 2-way 5-shot setting on PASCAL-$5^{i}$, \cite{tian2020aaai}'s setup is also adopted. The main difference lies in the testing: a query in~\cite{wang2019iccv,liu2020eccv}'s setup can contain 1 or 2 object classes but in~\cite{tian2020aaai}'s setup must contain both classes; 2) in COCO-$20^{i}$, there exist two data splits (A \& B) in the literature~\cite{nguyen2019iccv}. We follow ~\cite{wang2019iccv,liu2020eccv} to report results on split-A.

\para{Evaluation protocol.} We adopt the commonly used mIoU for performance evaluation{~\cite{shaban2017bmvc,zhang2019cvpr,wang2019iccv,tian2020pami}}: for a class $l$, the Intersection over
Union (IoU$_l$) is computed as $\frac{tp_l}{tp_l + fp_l + fn_l}$, where $tp_l$, $fp_l$, and $fn_l$ is the number of true positives, false positives and false negatives over the test set, respectively. mIoU is the average IoU over all classes.
In the multi-way setting, assuming a query contains an object class, some pixels of this class are mistakenly predicted as other classes (not background) which do not appear in this query, these pixels should be considered as both false negatives of this class and false positives of other classes. Yet, in the evaluation of~\cite{wang2019iccv,liu2020eccv}, they were ignored in the computation of false positives over other classes. We have communicated this with the authors of~\cite{wang2019iccv} and agreed that an appropriate metric should include this computation. We denote by \emph{mIoU*} the new metric as our default to distinguish from the \emph{mIoU} used in~\cite{wang2019iccv,liu2020eccv}. For a fair comparison, we report results on both. Notice there is no difference between \textbf{mIoU} and \textbf{mIoU*} in the 1-way setting, or in the multi-way setting when all targeted classes appear in the query.



\begin{figure*}[ht]
    \centering
    \includegraphics[scale=0.55]{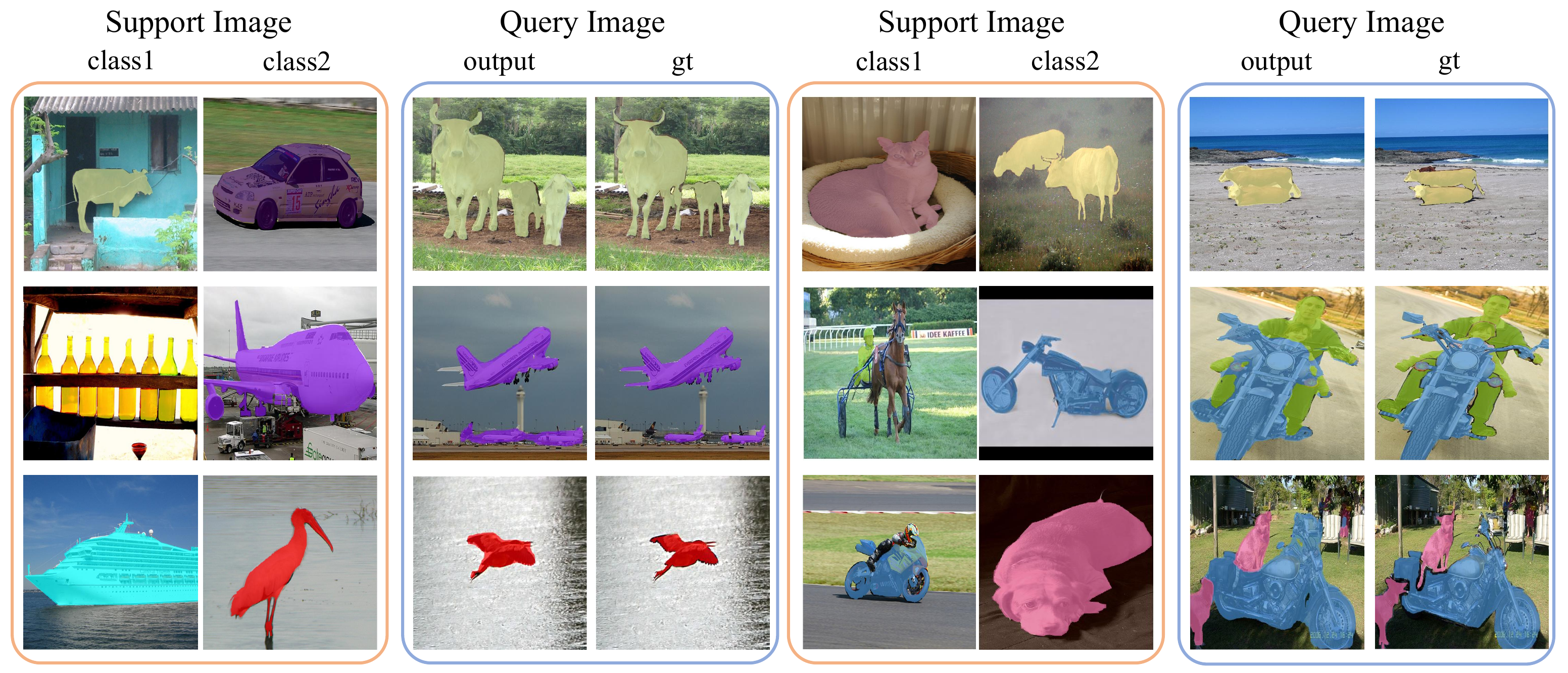}
    \caption{Qualitative results in 2-way 1-shot segmentation on PASCAL-$5^i$. Our outputs are close to the ground truth (gt). }
    \label{fig:result}
\end{figure*}

\subsection{Results on PASCAL-$5^{i}$}
\begin{table*}[t]
\caption{mIoU*/mIoU on PASCAL-$5^i$.
The mIoU results of \cite{wang2019iccv,liu2020eccv} and mIoU* results of \cite{Liu2020DynamicEN} were reported in their original papers while the mIoU* results of \cite{wang2019iccv,liu2020eccv} and mIoU results of \cite{Liu2020DynamicEN} were obtained by us re-testing their published models.  \miaojing{all methods employ the ResNet-50 backbone.}
\textbf{Params:} number of learnable parameters.
} 
\label{tab:voc}
	\centering
\footnotesize
\begin{tabular}{ccc|c|c|c|c|c|c|c|c|c}
\toprule
\multirow{2}{*}{\textsc{Method}}&\multirow{2}{*}{Parmas}& \multicolumn{5}{c}{2-way 1-shot}&  \multicolumn{5}{c}{2-way 5 -shot}\\
\cmidrule(r){3-7} \cmidrule(r){8-12}
 && fold-0 & fold-1 & fold-2 & fold-3 & Mean & fold-0 & fold-1 & fold-2 & fold-3 & Mean \\
\hline
PANet\cite{wang2019iccv}&23.5M &38.9/42.8&51.5/56.3&43.4/48.7&41.6/45.5&43.9/48.3&48.6/54.7&60.0/64.8&52.3/57.6&49.9/54.9&52.7/58.0\\
PPNet(w/o $U$)&23.5M&40.9/45.6&53.0/58.0&45.8/51.7&41.4/45.7&45.3/50.2&49.0/55.3&61.7/66.4&59.5/63.8&51.3/\textcolor{blue}{56.9}&55.4/60.6\\
PPNet\cite{liu2020eccv}&31.5M& 41.6/\textcolor{blue}{47.4} &53.5/58.3 &47.7/\textcolor{blue}{52.7} & 43.4/48.2&46.5/51.7 &\textcolor{blue}{49.4}/\textcolor{blue}{55.5}& \textcolor{blue}{62.2}/\textcolor{blue}{67.3}&\textcolor{red}{59.9}/\textcolor{red}{64.4} &\textcolor{blue}{52.5}/\textcolor{red}{58.0} &\textcolor{blue}{56.0}/\textcolor{red}{61.3}\\
DENet\cite{Liu2020DynamicEN}&13.9M&\textcolor{blue}{44.3}/46.7&\textcolor{blue}{61.9}/\textcolor{blue}{63.3}&\textcolor{red}{55.7}/\textcolor{red}{57.9}&\textcolor{blue}{46.7}/\textcolor{blue}{49.2}&\textcolor{blue}{52.2}/\textcolor{blue}{53.3}&45.5/52.2&62.1/67.2&\textcolor{blue}{58.7}/\textcolor{blue}{62.7}&48.1/52.7&53.6/58.7\\
MFNet&9.4M&\textcolor{red}{53.8}/\textcolor{red}{55.1} &  \textcolor{red}{66.8}/\textcolor{red}{68.0}&\textcolor{blue}{49.2}/50.5 &\textcolor{red}{48.2}/\textcolor{red}{50.1}&\textcolor{red}{54.5}/\textcolor{red}{55.9} &\textcolor{red}{58.5}/\textcolor{red}{60.3}&\textcolor{red}{70.0}/\textcolor{red}{70.9}&55.6/57.0&\textcolor{red}{54.6}/55.4&\textcolor{red}{59.7}/\textcolor{blue}{60.9}\\
\bottomrule
\end{tabular}
\end{table*}
\begin{table}[t]
\caption{Performance on PASCAL-$5^{2}$ with the setup of \cite{tian2020aaai}.}
    \label{tab:aaai}
    \centering
    \begin{tabular}{cc}
    \toprule
    \textsc{Method}& 2-way 5-shot\\
    \hline
         SG-One\cite{zhang2020tc} &39.4  \\
         PANet\cite{wang2019iccv} &41.3\\
         PLSEG\cite{dong2018bmvc} &42.6\\
         MetaSeg\cite{tian2020aaai}&43.3\\
         \hline
         MFNet &\textbf{47.8}\\
         \bottomrule
    \end{tabular}

      \vspace{-0.3cm}
\end{table}

\begin{table}[t]
\caption{Performance on PASCAL-$5^i$ in the 1-way 1-shot and 5-shot setting. Results of~\cite{wang2019iccv,zhang2019iccv} were copied from\cite{liu2020eccv}. \miaojing{all methods employ the ResNet-50 backbone.}    }
    \label{tab:1way}
    \centering
    \begin{tabular}{ccc}
    \toprule
         \textsc{Method}& 1-way 1-shot & 1-way 5-shot\\
    \midrule
     PPNet\cite{liu2020eccv}  & 52.8 &63.0\\
    RPMMs\cite{Yang2020PrototypeMM} &56.3 &  57.3\\
    PFENet\cite{tian2020pami}&{60.8}&61.9\\
    \miaojing{DENet\cite{Liu2020DynamicEN}}&60.1&60.5\\
    SCL \cite{Zhang2021SelfGuidedAC}&61.8&62.9\\
   {ASR}\cite{Liu2021AntialiasingSR}&58.2&61.0\\
    ASGNet \cite{Li2021AdaptivePL} & 59.3&63.9\\
    {MiningFSS} \cite{yang2021mining}&63.6&66.8\\
    CMN \cite{Xie2021ICCV}&62.8&63.7\\
    \miaojing{HSNet \cite{min2021hypercorrelation}}&64.0&\textcolor{red}{69.5}\\
      CWT\cite{lu2021iccv} &56.4&63.7\\
    CyCTR \cite{zhang2021few} &\textcolor{blue}{64.0}&\textcolor{blue}{67.5}\\
    \hline
    MFNet & 60.9 &  62.3 \\
    MFNet+ SA in~\cite{zhang2021few}&\textcolor{red}{64.5}&66.0\\
        \bottomrule
    \end{tabular}

\end{table}

\medskip

\para{\textbf{Multi-way: comparison to SOTA} in both the standard setup~\cite{wang2019iccv,liu2020eccv} and \cite{tian2020aaai}'s setup (see Sec.~\ref{sec:implement}}).

\noindent  \emph{\textbf{\cite{wang2019iccv,liu2020eccv}'s setup.}} \cite{wang2019iccv,liu2020eccv,Liu2020DynamicEN} offer the multi-way results, which are currently the state of the art in multi-class FSS. We compare with them on both mIoU* and mIoU using the same backbone ResNet-50 in Table~\ref{tab:voc}.
In the 2-way 1-shot, for mIoU*, MFNet is clearly better than{~\cite{wang2019iccv,liu2020eccv,Liu2020DynamicEN}} with a significant improvement of +$8.0\%$ over~\cite{liu2020eccv}  {and +$2.3\%$ over~\cite{Liu2020DynamicEN} on average}. Similar observations can be found on the mIoU metric.
As discussed in Sec.~\ref{sec:implement}, mIoU* is more appropriate than mIoU when evaluating queries in the multi-way setting. The large improvement by MFNet on mIoU* indicates that it is able to correctly predict the labels of both foreground and background in these queries. In the {2-way 5-shot} setting: MFNet is higher than~\cite{liu2020eccv} by +3.7\%  and +6.1\% than ~\cite{Liu2020DynamicEN} on the mean mIoU*. We also mark that~\cite{liu2020eccv} has utilized unlabeled data to boost the performance while we don't; without using the unlabeled data (PPNet w/o U), the results of ~\cite{liu2020eccv} will be lower.

We did not elaborate the performance on every fold. In general, compared to others, MFNet performs the best on every fold except for fold-2, which has also disadvantaged MFNet on the mean mIoU*. \miaojing{We investigated the performance of MFNet on individual classes within fold-2 and found out that MFNet produces rather low mIoU* on the “dining table” and “person” classes, \ie 12.9 and 9.1 respectively. The object appearances of these two classes vary drastically over images, which can be very challenging to detect.
We suggest that extracting only one prototype from such a class may not be sufficient to represent the class, we plan to explore it in-depth in future. Despite this, we emphasize the mean mIoU* over all folds is indeed more important to measure the overall performance of few-shot segmentation on PASCAL-5$^i$ (see also the discussion in~\cite{wang2019iccv,liu2020eccv}). Our MFNet improves the state of the art on the mean mIoU* significantly.
}

Table ~\ref{tab:voc} also shows that we achieve the best results with lowest number of learnable parameters in our model. Fig~\ref{fig:result} offers some visualization results.

\noindent  \emph{\textbf{\cite{tian2020aaai}'s setup.}}
\cite{tian2020aaai} has a different setup from~\cite{wang2019iccv,liu2020eccv}: images in PASCAL-$5^{i}$ which contain person and another held-out class are sampled to construct support and query sets for new tasks. \cite{tian2020aaai} has only reported the 2-way 5-shot results so we follow it and report our results in Table~\ref{tab:aaai}. In the {2-way 5-shot} setting of~\cite{tian2020aaai}, every query contains two object classes, thus mIoU* is the same to mIoU. Ours has mIoU* 47.8 which significantly improves~\cite{tian2020aaai} by 4.5\%. Results of~\cite{zhang2020tc,wang2019iccv,dong2018bmvc} were copied from~\cite{tian2020aaai} for comparison.
\medskip


\para{\textbf{Single-way: comparison to SOTA.}} MFNet is mainly proposed for the multi-way setting, but we can degrade it into the {1-way} setting.
The mean value of mIoU* over four folds are offered in Table~\ref{tab:1way} (mIoU* is the same to mIoU in the 1-way setting). Comparisons are among the recent state of the arts~\cite{liu2020eccv,tian2020pami,Li2021AdaptivePL,Yang2020PrototypeMM,Zhang2021SelfGuidedAC,Xie2021ICCV,lu2021iccv,yang2021mining,zhang2021few,min2021hypercorrelation,Liu2020DynamicEN,Liu2021AntialiasingSR}. All methods employ the same ResNet-50 as the backbone.
Our degraded MFNet in the 1-way setting achieves a competitive result with a relatively simple architecture (\miaojing{for instance, our result is on par with that in many representative works~\cite{Li2021AdaptivePL,Liu2021AntialiasingSR,Zhang2021SelfGuidedAC,lu2021iccv} published in 2021}). We can further combine it with the previous best~\cite{zhang2021few}: we use the multi-head transformer-based self-alignment (SA) block in ~\cite{zhang2021few} to enrich the multi-scale query representations in MFNet. This can boost the performance of MFNet to the state of the art result (\eg the highest mIoU 64.5 in 1-way 1-shot) as shown in Table~\ref{tab:1way}: MFNet + SA. The SA block in \cite{zhang2021few} is basically a pixel-level transformer block, which improves the robustness of feature representations while also increases the computational cost.

\miaojing{Overall, we argue that the study of FSS in the single-way and multi-way settings faces different challenges. Many FSS techniques proposed in the single-way setting may not be applicable or easily extendable to the multi-way setting. The primary focus in our paper is to propose techniques such as multi-level attention and pixel-wise metric learning that work well in the multi-way setting.}

\begin{table}[t]
\caption{Ablation study on the feature fusion. } 
\label{tab:fusion}
\center
\begin{tabular}{cccccc}
\toprule
\multirow{2}{*}{Fusion}& \multicolumn{5}{c}{2-way 1-shot} \\
\cmidrule(r){2-6}
 & fold-0 & fold-1 & fold-2 & fold-3 & Mean \\
\midrule
Baseline &  40.3 & 52.8 & 40.6 & 38.7 & 43.1 \\
\midrule
A+A & 48.9& 62.8& 45.1&45.0&50.5\\
C+C & 52.4& 65.5& 47.0 &46.8 &52.9 \\
C+A & 52.1& 65.4&46.8&45.8&52.5 \\
\textbf{MFNet (A+C)} & \textbf{53.8} & \textbf{66.8} & \textbf{49.2} & \textbf{48.2} & \textbf{54.5}\\
\bottomrule
\end{tabular}
\end{table}

\mshi{\para{\textbf{Transferring from single-way to multi-way.}}
In the multi-way setting, we have compared to several recent methods~\cite{wang2019iccv,liu2020eccv,Liu2020DynamicEN,tian2020aaai} who solve the multi-class segmentation as a one-to-many problem via explicitly learned class prototypes/classifiers. Our method outperforms them significantly!
Unlike these methods, MFNet realizes the multi-class few-shot segmentation via a pixels-to-pixels DNN. The pixels-to-pixels DNN has been utilized in recent FSS works~\cite{zhang2019cvpr,zhang2019iccv,tian2020pami,liu2020cvpr,Zhang2021SelfGuidedAC,lu2021iccv,zhang2021few} but only in the single-way setting. We are the first to propose a novel multi-way encoding \& decoding scheme. 
To justify our design, we compare it to two SOTA methods~\cite{zhang2021few,tian2020pami} by applying a naive transfer on them from the 1-way to $N$-way: we forward the binary segmentation pipeline learned in these works for $N$ times with each one being responsible for predicting one of the $N$ classes. The predicted label for a pixel can be as either background or a specific class in each time. Overall, we take the class label (including background) at each pixel with the maximum probability over $N$ times. We report in Table~\ref{tab:1to2} the results in 2-way 1-shot and 5-shot. It can be seen that our method outperforms \cite{zhang2021few,tian2020pami} significantly. Moreover, we also provide the network inference time for one image in 2-way 1-shot: \cite{zhang2021few,tian2020pami} spent about twice the time as we do (they have to be forwarded twice). In summary, in the multi-class scenario, our method is apparently much more efficient and effective than sequentially segmenting each class by employing a single-class few-shot segmentation model.  }
\begin{table}[t]
\caption{Comparison to \cite{tian2020pami,zhang2021few} on PASCAL-5$^i$ in 2-way 1-shot and 5-shot. The network inference time for one image was reported in 2-way 1-shot. We use * to denote that results of \cite{tian2020pami,zhang2021few} were obtained by our implementation.}
    \label{tab:1to2}
    \centering
    \begin{tabular}{cccc}
    \toprule
         \textsc{Method}&inference time& 2-way 1-shot & 2-way 5-shot \\
    \midrule
    PFENet*\cite{tian2020pami}&0.096s &42.3&48.4\\
     CyCTR*\cite{zhang2021few}&0.118s&47.6&54.1\\
    \hline
     MFNet&0.057s&54.5&59.7\\
        \bottomrule
    \end{tabular}
\end{table}

\para{Ablation study.} We focus on details of the multi-way encoding \& decoding, multi-level attention, and  pixel-wise metric learning in the 2-way setting.

\noindent \emph{\textbf{Multi-way encoding \& decoding.}}  The proposed architecture encodes the multi-way input into one feature and decodes multi-class segmentation on it. One baseline is devised: we construct multiple binary encoding \& decoding branches with a shared backbone. Each branch encodes the multi-scale query features with the prototype of one class and decodes binary segmentation result for this class. The multi-class segmentation result is obtained by taking the class (including background) with the maximum probability over multiple branches at each pixel. The proposed pixel-wise metric learning is not applied in this baseline. 
Results in Table~\ref{tab:fusion} show that the baseline performs significantly lower than our MFNet, which demonstrates the effectiveness of our method.

Next, we validates our choice of {Add} (A) and {Concat} (C) in the multi-way encoding process. Referring to Sec.~\ref{Sec:MED}, we first \emph{add} each expanded class prototype to the query feature of certain scale and later \emph{concatenate} the fused features over multiple classes. We enumerate four possible combinations of A and C in the two steps in Table~\ref{tab:fusion}. It clearly shows that A + C works the best with mIoU* 54.5. C + C performs the second (52.9) while A + A the worst (50.5). The results are consistent with the analysis offered in Sec.~\ref{Sec:MED}.

\noindent \emph{\textbf{Attention A$^\text{S}$ for support feature modulation.}}
A$^\text{S}$ is proposed for the multi-shot support feature modulation to effectively extract one prototype from multiple support features per class. Hence, we run the ablation study in the 2-way 5-shot setting. \mshi{We first compare with the result in the 2-way 1-shot setting (Table~\ref{tab:fusion}: bottom) where A$^\text{S}$ does not apply: the improvement of MFNet from 2-way 1-shot to 2-way 5-shot is 5.2 in mIoU*.} Next, we ablate A$^\text{S}$ in the 2-way 5-shot: we run MFNet with and without A$^\text{S}$ (MFNet w/ A$^\text{S}$ \vs MFNet w/o A$^\text{S}$) and report the results in Table~\ref{tab:attention}, where clear improvement can be observed by adding A$^\text{S}$ in MFNet. Besides the proposed relational attention, we tried to adapt two attention mechanisms from~\cite{zhang2019cvpr} and~\cite{azad2021wacv} to combine multiple support features: a query feature is firstly concatenated with multiple support features, respectively; these features can be combined through weights of softmaxed self-attention (MFNet w/ A$^\text{S}$-v1) or through Bi-ConvLSTM (MFNet w/ A$^\text{S}$-v2) to produce one feature for each query scale. The proposed attention for multi-scale combination is still applied afterwards.
Results are in Table~\ref{tab:attention}: we did not observe meaningful improvement by utilizing A$^\text{S}$-v1/A$^\text{S}$-v2. In contrast, the proposed A$^\text{S}$ improves the mean mIoU* from 58.5 to 59.7.

\begin{table}[t]
\caption{Ablation study on attention for support feature modulation. } 
\label{tab:attention}
\center
\small
\begin{tabular}{cccccc}
\toprule
\multirow{2}{*}{Attention}& \multicolumn{5}{c}{2-way 5-shot} \\
\cmidrule(r){2-6}
 & fold-0 & fold-1 & fold-2 & fold-3 & Mean \\
\hline
w/o A$^\text{S}$ &57.6&68.8&53.5&54.2&58.5\\
w/ A$^\text{S}$-v1&58.1&68.8&53.8&54.4&58.7\\
w/ A$^\text{S}$-v2&57.8&68.7&53.5&53.8&58.5\\
\textbf{ MFNet (w/ A$^\text{S}$)}&\textbf{58.5}&\textbf{70.0}&\textbf{55.6}&\textbf{54.6}&\textbf{59.7}\\
\bottomrule
\end{tabular}
\end{table}

\noindent \emph{\textbf{Attention A$^\text{M}$ for multi-scale combination.}}
The proposed idea, denoted by MFNet w/ A$^\text{M}$, is compared to that without using multi-scale nor attention (MFNet w/o A$^\text{M}$), where one could clearly see the improvement in Table~\ref{tab:selfattention}.
Besides using attended weights for multi-scale combination, we also tried two alternative forms: one follows the FEM module in \cite{tian2020pami} to hierarchically concatenate the multi-scale features before decoding;
another follows the ASPP module in \cite{chen2017arxiv} to implement the multi-scale features in the decoder.
We denote them by MFNet w/ A$^\text{M}$-v1 and MFNet w/ A$^\text{M}$-v2 in Table~\ref{tab:selfattention}, respectively. Our MFNet w/ A$^\text{M}$ performs clearly better than the two variants.
\begin{table}[t]
\caption{Ablation study on attention for multi-scale combination. } 
\label{tab:selfattention}
\center
\small
\begin{tabular}{cccccc}
\toprule
\multirow{2}{*}{Attention}& \multicolumn{5}{c}{2-way 1-shot} \\
\cmidrule(r){2-6}
 & fold-0 & fold-1 & fold-2 & fold-3 & Mean \\
\hline
w/o A$^\text{M}$ &51.5&65.6&47.9&45.9&52.7\\
w/ A$^\text{M}$-v1&53.4&66.1&48.7&47.5&53.9\\
w/ A$^\text{M}$-v2 &52.4&66.0&47.6&46.9&53.2\\
\textbf{ MFNet (w/ A$^\text{M}$)}&\textbf{53.8}&\textbf{66.8}&\textbf{49.2}&\textbf{48.2}&\textbf{54.5}\\
\bottomrule
\end{tabular}
\end{table}

\begin{table}[t]
\caption{Ablation study on the PML-triplet formation. }
\label{tab:pml}
\center
\renewcommand\tabcolsep{3pt}
\small
\begin{tabular}{cccccc}
\toprule
\multirow{2}{*}{PML}& \multicolumn{5}{c}{2-way 1-shot} \\
\cmidrule(r){2-6}
 & fold-0 & fold-1 & fold-2 & fold-3 & Mean \\
 \hline
w/o PML & 52.8&65.0&47.7&46.6&53.0\\
w/ PML-rnd&53.2&66.1&48.7&47.3&53.8\\
w/ PML-fea& 53.0&65.2&48.9&47.1&53.6\\
\textbf{MFNet (w/ PML-spat)} & \textbf{53.8} & \textbf{66.8}& \textbf{49.2}& \textbf{48.2}& \textbf{54.5}\\
\bottomrule
\end{tabular}

   \vspace{-0.3cm}
\end{table}


 \begin{figure*}[t]
     \centering
    \includegraphics[scale=0.6]{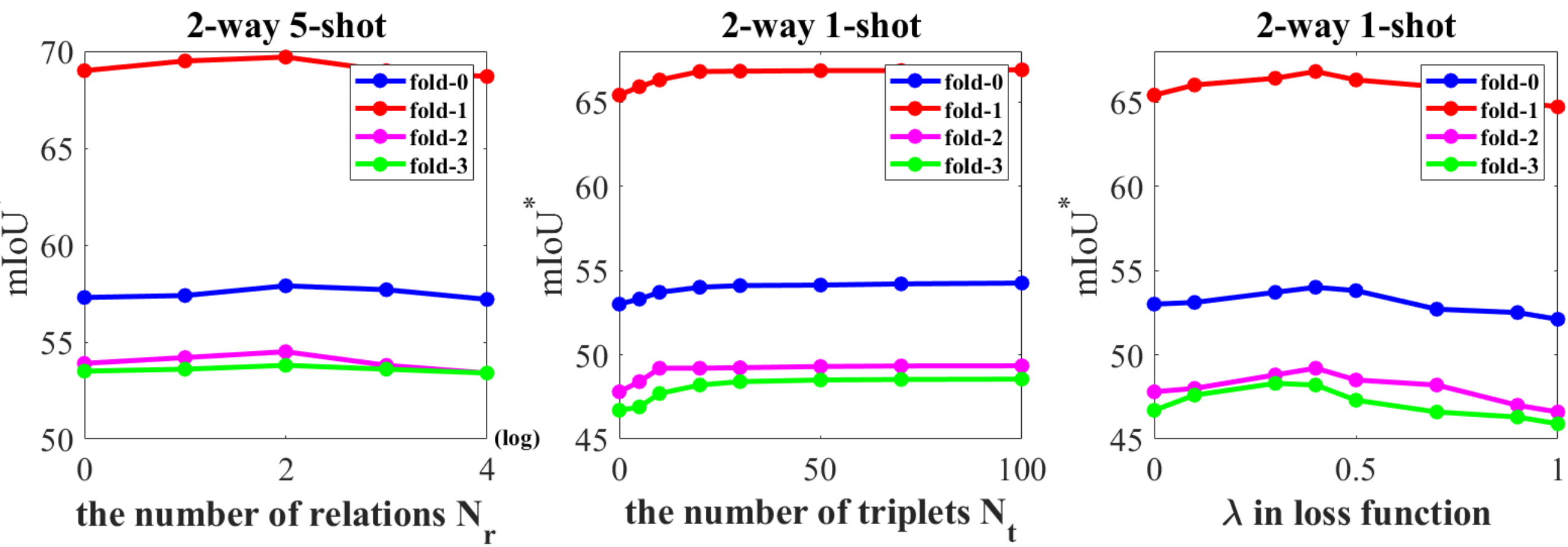}
     \caption{Parameter variations of $N_r$, $N_t$ and $\lambda$.}
\label{fig:pparameter}
\end{figure*}


\noindent \emph{\textbf{PML-triplet formation.}}
Computing high-dimensional feature distances of pixels is too expensive for triplet formation, especially for online updating. We introduce a weighted random selection strategy using pixels' spatial distances and predicted labels (PML-spat) as proxies.
To compare with our method, we offer two variants: random selection (PML-rnd) and feature distance-based selection (PML-fea). Both are still confined within the same pools of hard negatives, hard positives, and anchors to PML-spat for the sake of efficiency. Results are shown in Table~\ref{tab:pml}: 1) all the three variants of PML are better than not using it (MFNet w/o PML); 2) our PML-spat performs the best over PML-rnd and PML-fea. Notice PML-rnd and PML-fea can perform even worse if they do not use data pools proposed for PML-spat. In PML-spat, the spatial distance between pixels can be computed offline, it is not only effective but also very fast.

\medskip

\begin{table}[t]
\caption{ Ablation study on the backbone.}
    \label{tab:backbone}
    \centering
    \begin{tabular}{ccc}
    \toprule
         \textsc{Method}& 2-way 1-shot & 2-way 5-shot\\
    \midrule
     MFNet (VGG16) & 52.7  & 57.6 \\
      MFNet (ResNet50) &\textbf{54.5} &\textbf{59.7}\\
        \bottomrule
    \end{tabular}
\end{table}

\begin{figure*}[ht]
    \centering
    \includegraphics[scale=0.65]{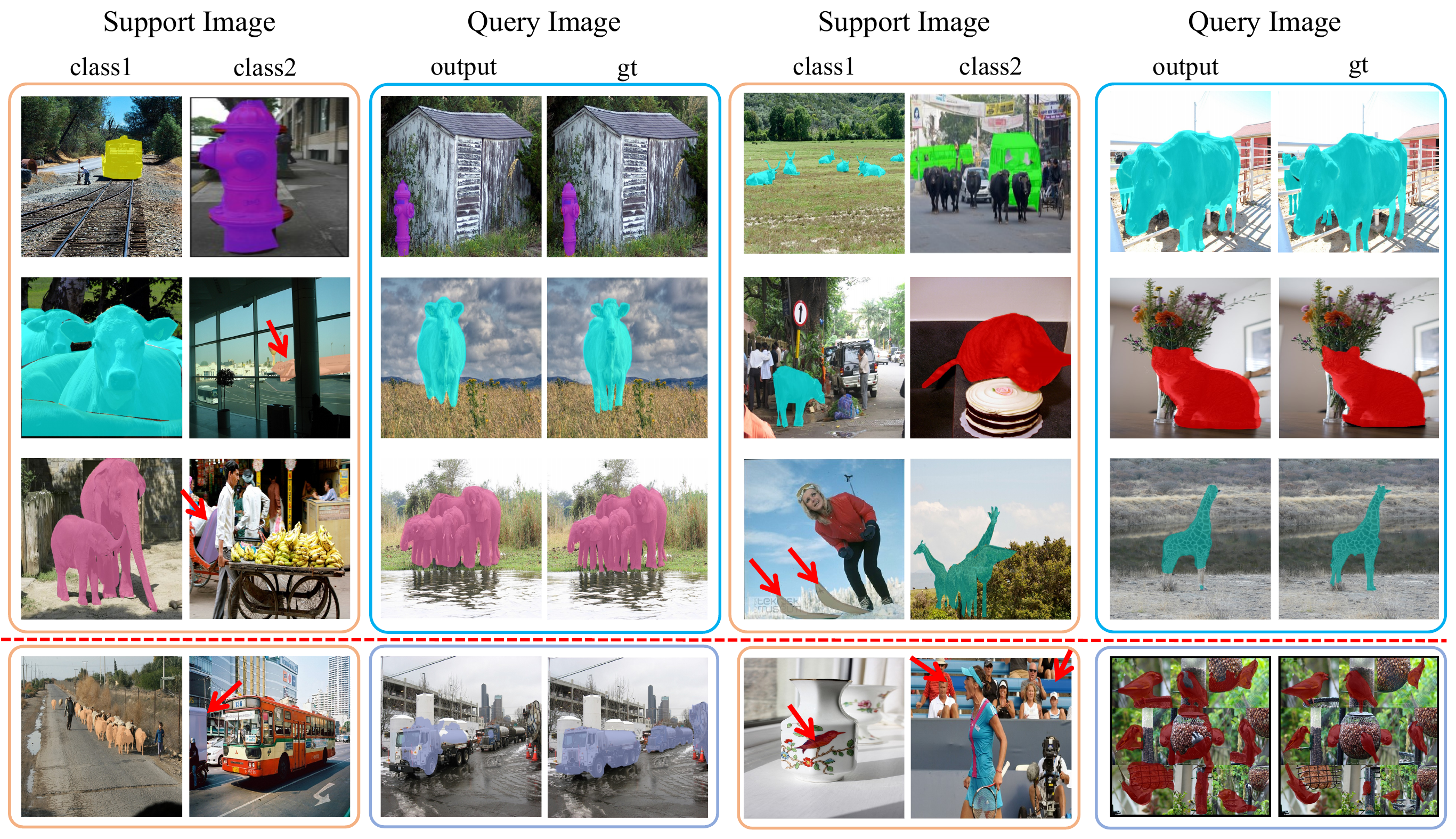}
    \caption{Qualitative results in the 2-way 1-shot segmentation on COCO-$20^i$. \miaojing{Images below the red line are false cases. Tiny mask labels have been noted by red arrows for clarity.} }
    \label{fig:result-coco}
\end{figure*}

\begin{table*}[t]
\caption{mIoU*/mIoU on COCO-$20^{i}$. The mIoU results of \cite{wang2019iccv,liu2020eccv} were reported in their original papers while the mIoU* results of them were obtained by us re-testing their published models.
* Results of \cite{Liu2020DynamicEN} were obtained by our implementation.
\miaojing{all methods employ the ResNet-50 backbone.}
\textbf{Params:} number of learnable parameters. }
    \label{tab:coco}
	\centering
\footnotesize
\begin{tabular}{cc|c|c|c|c|c|c|c|c|c}
\toprule
\multirow{2}{*}{\textsc{Method}}& \multicolumn{5}{c}{2-way 1-shot}&  \multicolumn{5}{c}{5-way 1-shot}\\
\cmidrule(r){2-6} \cmidrule(r){7-11}
 & fold-0 & fold-1 & fold-2 & fold-3 & Mean & fold-0 & fold-1 & fold-2 & fold-3 & Mean \\
 \hline
 PANet\cite{wang2019iccv}& 25.7/31.9&16.1/21.5&16.2/21.3&13.8/16.4&18.0/22.8&22.1/27.2&16.0/21.5&14.1/19.7&11.9/15.4&16.0/20.9\\
 PPNet(w/o $U$)&29.0/33.9&19.4/24.0&16.5/22.8&14.2/17.6&19.8/24.6&24.3/29.1&16.8/22.3&14.3/\textcolor{blue}{21.1}&12.8/16.4&17.1/22.2\\
 PPNet\cite{liu2020eccv}&29.8/34.2&19.7/24.2&\textcolor{blue}{17.0}/\textcolor{red}{23.4}&15.1/19.1&20.4/25.2&25.6/30.8&17.3/\textcolor{blue}{23.0}&\textcolor{red}{15.5}/\textcolor{red}{21.3}&13.4/17.9&18.0/23.3\\
 DENet*\cite{Liu2020DynamicEN}&\textcolor{blue}{30.3}/\textcolor{blue}{35.2}&\textcolor{blue}{20.5}/\textcolor{blue}{25.1}&16.7/\textcolor{blue}{23.0}&\textcolor{blue}{15.3}/\textcolor{red}{21.1}&\textcolor{blue}{20.7}/\textcolor{blue}{26.1}&\textcolor{blue}{27.0}/\textcolor{blue}{32.7}&\textcolor{blue}{17.9}/22.4&14.6/19.9&\textcolor{blue}{14.0}/\textcolor{blue}{20.5}&\textcolor{blue}{18.4}/\textcolor{blue}{23.9}\\
 MFNet& \textcolor{red}{35.3}/\textcolor{red}{39.0}&\textcolor{red}{24.0}/\textcolor{red}{28.9}&\textcolor{red}{18.4}/21.5&\textcolor{red}{18.7}/\textcolor{blue}{22.9}&\textcolor{red}{24.1}/\textcolor{red}{28.1}&\textcolor{red}{29.7}/\textcolor{red}{35.5}&\textcolor{red}{21.2}/\textcolor{red}{26.7}&\textcolor{blue}{15.4}/20.0&\textcolor{red}{17.1}/\textcolor{red}{21.1}&\textcolor{red}{20.9}/\textcolor{red}{25.8}\\
\bottomrule
    \end{tabular}

\end{table*}
\begin{table}[t]
\caption{Performance on COCO-$20^i$ in the 1-way 1-shot and 5-shot setting.
* Results of ~\cite{zhang2021few} were obtained by our implementation.
\miaojing{all methods employ the ResNet-50 backbone.}  }
    \label{tab:coco-1way}
    \centering
    \begin{tabular}{ccc}
    \toprule
         \textsc{Method}& 1-way 1-shot & 1-way 5-shot\\
    \midrule
     PANet\cite{wang2019iccv} & 23.0 & 33.8\\
      PPNet(w/o $U$) &25.7&36.2\\
     PPNet\cite{liu2020eccv}  & 27.2&36.7\\
     CWT\cite{lu2021iccv}&32.9&41.3\\
     MiningFSS\cite{yang2021mining}&35.1&\textcolor{blue}{41.4}\\
     CyCTR*\cite{zhang2021few}&\textcolor{blue}{35.7}&41.0\\
    \hline
    MFNet& 34.9 &  39.2 \\
    {MFNet + SA in~\cite{zhang2021few}}&\textcolor{red}{37.5} &\textcolor{red}{41.9} \\
        \bottomrule
    \end{tabular}
\end{table}

\para{\textbf{Parameter variations.}} We study the number of relational features $N_r$, number of triplets $N_t$, and $\lambda$ in the loss function. Results are shown in the 2-way 1-shot and 5-shot where appropriate.

\noindent  \emph{\textbf{Number of relational features $N_r$.}} In Fig.~\ref{fig:pparameter}: left we vary the number of relational features $N_r$ for each support feature in the attention scheme. Similar to~\cite{hu2018cvpr}, we vary $N_r$ among 1, 2, 4, 8, and 16. The best performance occurs at $N_r = 4$. Recalling our feature vector is of 256 dimensions, if $N_r$ is large, \eg 16, the channel of each $R_{nk}^r$ is only $\frac{256}{16}$ dimensions. This is too small for relational features, thus less meaningful.

\noindent \emph{\textbf{Number of triplets $N_t$.}} In Fig.~\ref{fig:pparameter}: middle we vary the number of triplets $N_t$ for PML from 0 to 100. The results show that mIoU* increases quickly from 0 to 20 and slowly afterwards over the four folds. Meanwhile, the computational cost increases proportionally to $N_t$. We choose $N_t = 20$ as a good tradeoff between performance and efficiency. Benefit of further increasing $N_t$ is not worth of the extra cost.

\noindent {\emph{\textbf{$\lambda$ in loss function.}} The triplet loss in PML serves as an auxiliary loss of our segmentation task. We combine it with the focal segmentation loss using a parameter of $\lambda$ in (\ref{eq:loss}). In Fig.~\ref{fig:pparameter}: right we vary $\lambda$ within a range of [0,1] over the four folds of  PASCAL-5$^i$ in the 2-way 1-shot setting. Results show that the best performance occurs at $\lambda = 0.4$ overall. This is also our default setting.}

\medskip

\para{\textbf{Backbone choice.}}
\miaojing{ResNet-50 is a commonly used backbone in FSS, which is also our default backbone. Here, we provide the experiment using an alternative backbone: the VGG net~\cite{simonyan2015iclr}. Results in Table~\ref{tab:backbone} show that MFNet (ResNet-50) performs better than MFNet (VGG) in both 2-way 1-shot and 5-shot. This is consistent to the results observed in~\cite{liu2020eccv,tian2020pami,Yang2020PrototypeMM} who have also reported results using both backbones.}

\subsection{Results on COCO-$20^{i}$}
\para{\textbf{Multi-way: comparison to SOTA.}} Following~\cite{wang2019iccv,liu2020eccv}, Table~\ref{tab:coco} presents the 2-way 1-shot and 5-way 1-shot results on COCO-20$^i$. Comparisons are still with~\cite{wang2019iccv,liu2020eccv,Liu2020DynamicEN} using the same data split and same backbone ResNet-50. ~\cite{Liu2020DynamicEN} has originally used a different data split to report results on COCO-20$^i$  (Sec.~\ref{sec:implement}). Its results in Table~\ref{tab:coco} were obtained by ourselves. In 2-way 1-shot, MFNet yields mean mIoU* 24.1 and mean mIoU 28.1, which is +3.7\% and +2.9\% over PPNet~\cite{liu2020eccv}, +4.3\% and +3.5\% over PPNet w/o U~\cite{liu2020eccv}, +3.4\% and +2.0\% over DENet~\cite{Liu2020DynamicEN}, respectively. Similar observation goes to 5-way 1-shot: we have mean mIoU*/mIoU as 20.9/25.8, 18.4/23.9, 18.0/23.3, 17.1/22.2, 16.0/20.9 for MFNet, {DENet~\cite{Liu2020DynamicEN}}, PPNet~\cite{liu2020eccv} and PPNet (w/o U)~\cite{liu2020eccv}, and PANet~\cite{wang2019iccv} correspondingly. MFNet is clearly the best.
An image in COCO-20$^i$ normally contain more object classes than in PASCAL-5$^i$, thus is more challenging for FSS.
{Fig.~\ref{fig:result-coco} shows qualitative results in the 2-way 1-shot setting on COCO-20$^i$. Some failure cases are presented in the figure.}

\medskip
\para{\textbf{Single-way: comparison to SOTA.}} Here we offer the results in the 1-way 1-shot and 1-way 5-shot.
Comparisons are made to~\cite{liu2020eccv,wang2019iccv,lu2021iccv,yang2021mining,zhang2021few} {in the same data split} and with the same backbone ResNet-50. ~\cite{zhang2021few} has originally used a different data split to report results on COCO-20$^i$ (Sec.~\ref{sec:implement}). Its results in Table~\ref{tab:coco-1way} were obtained by ourselves.
We also present the results of PPNet (w/o U) without using the unlabeled dataset~\cite{liu2020eccv}. Table~\ref{tab:coco-1way} shows that our MFNet performs very close to the previous best~\cite{zhang2021few} and~\cite{yang2021mining} in the 1-way 1-shot and 1-way 5-shot, respectively.
Similar to what we did on MFNet in Table~\ref{tab:1way}, we can add the SA block in \cite{zhang2021few} to MFNet and further improve its performance on COCO-20$^i$. Table~\ref{tab:coco-1way} shows that MFNet + SA produces the highest mIoU, 37.5 and 41.9, in the 1-way 1-shot and 5-shot, respectively.


\section{Conclusion}
In the study of few-shot learning, few-shot semantic segmentation has not been largely explored. Representative works are mostly restrictive to single-class segmentation; a few are extendable to multi-class segmentation yet with non-parametric metric learning modules mostly. \mshi{This work first introduces a novel multi-way encoding and decoding pipeline to tackle multi-class FSS. For better feature fusion, a multi-level attention mechanism is also presented: the attention for support feature modulation works in the multi-shot setting while the attention for multi-scale combination can particularly suit in the multi-class setting. Finally, to enhance the embedding space learning of query and support images, a pixel-wise triplet loss is devised on the query's pixel-level. It is effective for both single-class and multi-class FSS. }
We conduct extensive experiments on two standard benchmarks,
\ie PASCAL-$5^{i}$ and COCO-$20^{i}$. Results show that our method improves the state of the art by a large margin. Future work will be mainly focused on post-processing the output of our method for performance enhancement.

\section{Acknowledgment}
This work was supported by National Natural Science Foundation of China under Grant No. 61828602, 72171172 and 61903027, and Shanghai Municipal Science and Technology Major Project under Grant No. 2021SHZDZX0100.

\bibliographystyle{IEEEtran}

\end{document}